\documentclass{article} 
\usepackage{iclr2027_conference,times}

\usepackage{amsmath,amsfonts,bm}

\def\eqref#1{equation~\ref{#1}}

\def\1{\bm{1}}

\DeclareMathAlphabet{\mathsfit}{\encodingdefault}{\sfdefault}{m}{sl}
\SetMathAlphabet{\mathsfit}{bold}{\encodingdefault}{\sfdefault}{bx}{n}

\usepackage{graphicx}
\usepackage[dvipsnames]{xcolor}

\usepackage{booktabs}
\usepackage{adjustbox}
\usepackage{makecell}
\usepackage{multirow}
\usepackage{enumitem}
\usepackage{xspace}
\usepackage{tcolorbox}
\tcbuselibrary{skins}
\usepackage{hyperref}
\usepackage{cleveref}

\newcommand{\paragrapht}[1]{\hspace*{-\parindent}\textbf{#1}}

\definecolor{blue(pigment)}{rgb}{0.2, 0.2, 0.9}

\colorlet{myblue}{MidnightBlue!5}
\colorlet{myred}{RubineRed!5}

\usepackage{hyperref}  
\hypersetup{colorlinks=true, citecolor=blue(pigment), linkcolor=blue(pigment), urlcolor=blue(pigment)}
\usepackage{url}

\usepackage{wrapfig}   
\usepackage{caption}   

\title{Speculative Forcing for Efficient\\ Autoregressive Video Generation\vspace{-4mm}}

\author{Hyunwook Choi$^{1*}$, Seungryong Kim$^{1}$, Dongyoon Han$^{\dagger,2}$, Sangdoo Yun$^{\dagger,2}$ \\
$^{1}$KAIST AI, $^{2}$NAVER AI Lab \\
{\small \hypersetup{pdfborder={0 0 0}}\mdseries\href{https://speculative-forcing.github.io/}{\textcolor{blue}{\texttt{https://speculative-forcing.github.io/}}}}
}

\usepackage{lipsum}
\usepackage{booktabs}   
\usepackage{multirow}   
\usepackage{makecell}   
\usepackage[table]{xcolor} 
\usepackage{graphicx}
\usepackage{tikz}
\usepackage{pgfplots}
\pgfplotsset{compat=1.18}

\iclrfinalcopy 
\begin{document}

\maketitle
\NoHyper{\renewcommand{\thefootnote}{}\footnotetext{\hspace{-1.8em}$^{*}$Work done during internship at NAVER AI Lab. \quad $^{\dagger}$Corresponding authors.}}\endNoHyper
\lhead{Speculative Forcing for Efficient Autoregressive Video Generation}

\begin{figure}[htbp]
    \centering
   \vspace{-3em}
    \includegraphics[width=.925\linewidth]{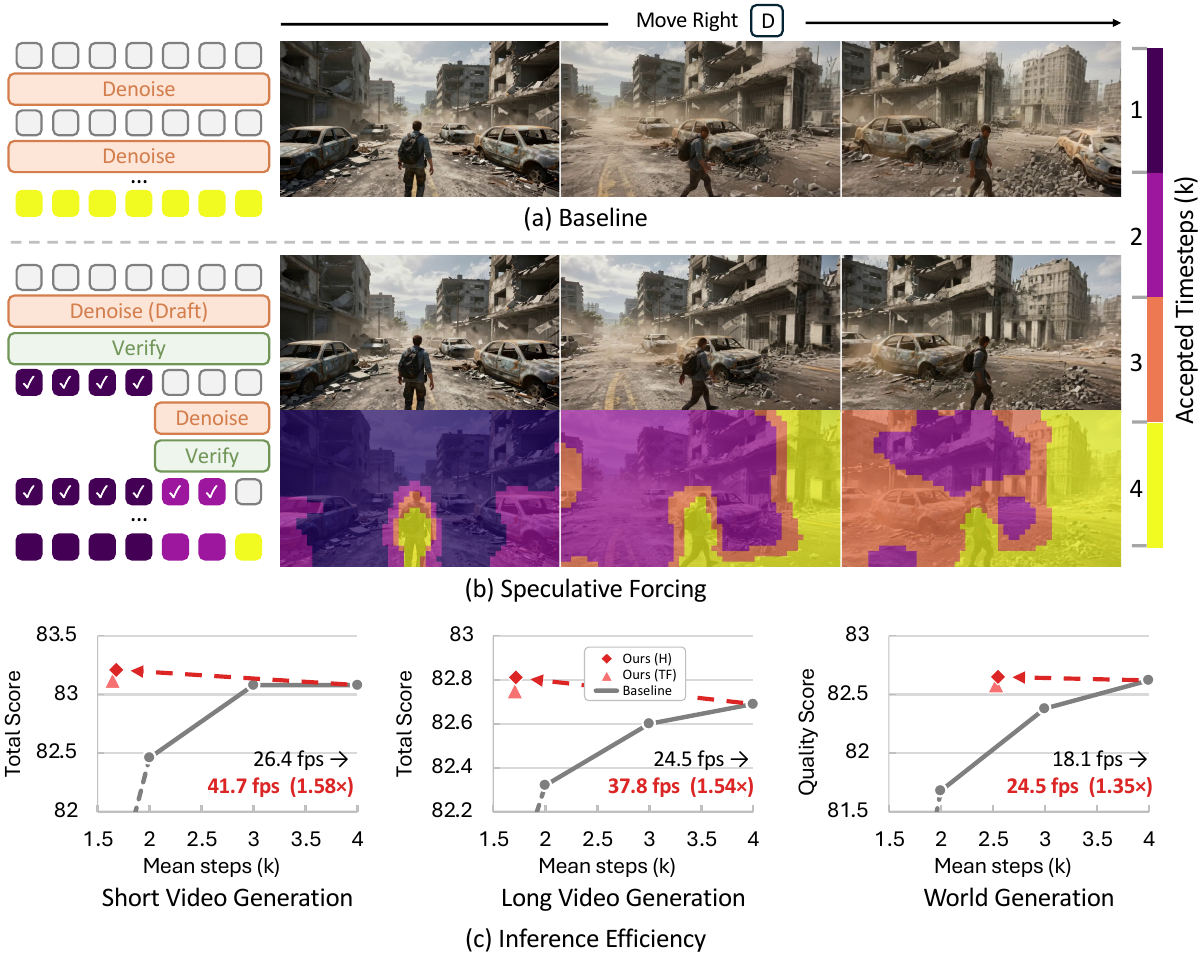}
    \vspace{-.5em}
    \caption{Compared with full-step sampling (a), Speculative Forcing treats intermediate token estimates as drafts, accepts those that have converged, and refines only the rest (b), achieving comparable quality with fewer denoising steps and higher throughput (c).}
    \vspace{-.5em}
\label{fig:teaser}
\end{figure}

\begin{abstract}
\vspace{-.5em}
Few-step autoregressive video generators enable streaming generation, yet they run every token through the final denoising step, regardless of how much refinement it needs. We find this uniform computation is often unnecessary: many tokens already closely match their eventual full-step outputs in the initial steps, while others require refinement until the end. We introduce \textbf{Speculative Forcing}, a framework that treats intermediate clean estimates as drafts and predicts when individual tokens can be safely accepted early. Once accepted, a token is frozen as its final output and undergoes no further denoising, while its cached keys and values remain available as context for the remaining tokens. We explore two alternative verifier designs: a training-free one based on similarity to clean content and stability across steps, and a learned one that reads the frozen generator's intermediate features to estimate each token's remaining gap to its full-step output. Across short-video, long-video, and world generation, Speculative Forcing preserves the quality of full-step sampling while substantially reducing computation. On LongLive, it matches the four-step generator with $\sim$1.7 denoising steps per token, running over $1.5\times$ faster at 41.7 and 37.8\,fps. On Lingbot-World 2.0, where camera motion continually changes the visible content, it maintains quality with $\sim$2.5 denoising steps per token and runs $1.35\times$ faster at 24.5\,fps.
\end{abstract}    
\section{Introduction}
\label{sec:intro}

Diffusion models have enabled high-quality image and video generation, with recent video models achieving realistic appearance, natural motion, and faithful prompt following~\citep{wan, hunyuanvideo, cogvideox, cosmos2.5, seedance2}. Autoregressive (AR) video generators extend these capabilities to streaming generation~\citep{causvid, selfforcing, longlive, rollingforcing, causalforcing, rewardforcing} and interactive world models~\citep{lingbotworld1, lingbotworld2, matrixgame2, matrixgame3, swm, alayaworld}. Within a video, however, different regions may require different amounts of refinement. Some may readily reproduce familiar content from the history, while others must form newly appearing structures or resolve fine details. This raises a fundamental question: \textit{Does every token need to be denoised until the final step?}

We study this question in distilled few-step AR video generators, whose intermediate predictions can already serve as plausible video outputs. Self Forcing~\citep{selfforcing} uses intermediate clean estimates as outputs for distribution-matching training~\citep{dmd, dmd2}, making them natural candidates for early commitment. We observe that many tokens already approximate the generator's own full-step output in the initial steps, while others continue changing until the last step (Figure~\ref{fig:diff_map}). These differences often follow intuitive patterns. Regions that reproduce previously observed content can settle early, while nearby regions continue refining appearance or motion. Similar variation arises in world models as actions change the observed content. These observations motivate a simple principle: stop denoising tokens once they have converged, and spend the remaining computation only where refinement is still needed. 

We introduce \textbf{Speculative Forcing}, a framework that predicts when each token's estimate has sufficiently converged, allowing individual tokens to be accepted at different steps (Figure~\ref{fig:teaser}). Each intermediate clean estimate serves as a draft of the generator's full-step output. A verifier determines whether to accept the draft or continue refining it. Once accepted, the estimate is fixed as the token's output and receives no further denoising updates, while retaining its cached keys and values as context for the remaining tokens. This requires predicting whether a token's current draft is already close to its full-step output without running the remaining denoising steps.

We address this challenge with two verifiers. Because drafts are clean latent estimates, similarity to previously generated content and changes between consecutive estimates of the same token provide evidence of convergence without additional training. Our training-free verifier uses similarity to clean history or reference content at the first step and stability between consecutive drafts at later steps. We also introduce a learned verifier for demanding settings such as world models, leveraging contextual cues encoded in the pretrained generator's intermediate features. A shallow head trained with the generator frozen predicts each token's gap to the full-step output, showing that the intermediate features contain cues about how much its predictions will change with further denoising. Both verifiers support token-wise acceptance without modifying the generator's weights.

We evaluate Speculative Forcing on LongLive~\citep{longlive} and the world model Lingbot-World 2.0~\citep{lingbotworld2}. On both short and long videos, our method maintains the quality of the four-step generator with approximately 1.7 denoising steps per token, running over $1.5\times$ faster at 41.7 and 37.8\,fps. On controllable world model rollouts, a harder setting where camera motion continually changes the visible content, it maintains quality with approximately 2.5 denoising steps per token, running $1.35\times$ faster at 24.5\,fps.

\section{Related Work}
\begin{figure}
\centering
\includegraphics[width=.95\linewidth]{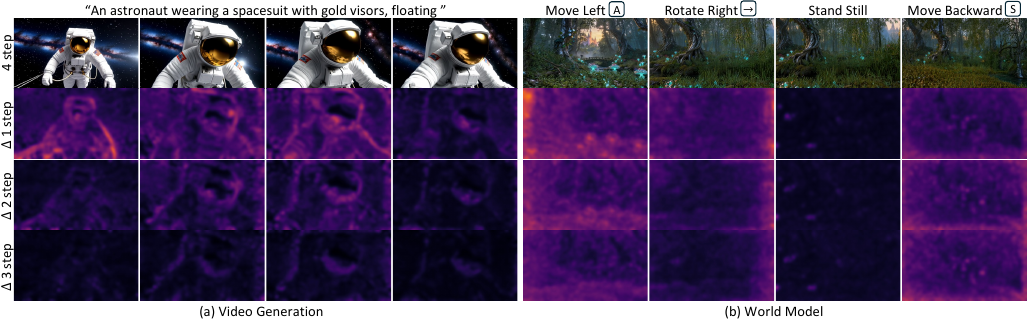}
\vspace{-.5em}
\caption{
\textbf{Token-wise convergence during denoising.}
We visualize latent differences between each chunk's $k$-step prediction and the baseline four-step output.
(a) In video generation, background and object regions that remain similar to history tend to converge early, while other regions continue to change. (b) In world generation, previously observed regions often converge early, whereas newly revealed regions tend to require further denoising.
}
\vspace{-1em}
\label{fig:diff_map}
\end{figure}

\paragrapht{Autoregressive Video Generation.} Autoregressive (AR) video diffusion models generate successive frames or chunks conditioned on preceding content~\citep{magi, skyreels, diffusionforcing}. Distribution matching distillation~\citep{dmd, dmd2} enables few-step generation by aligning the student's output distribution with a target distribution represented by a pretrained teacher. CausVid applies distribution matching to distill a bidirectional teacher into a causal generator~\citep{causvid}, while Self Forcing trains on autoregressive rollouts conditioned on the generator's own outputs~\citep{selfforcing}. Causal Forcing revisits this distillation for real-time interactive generation~\citep{causalforcing, causalforcingpp}. LongLive, Rolling Forcing, and Self-Forcing++ extend few-step generation to long streaming videos~\citep{longlive, rollingforcing, selfforcingpp}. The same few-step AR recipe underlies real-time interactive world models, which additionally condition generation on actions or camera trajectories~\citep{lingbotworld1, lingbotworld2, matrixgame2, matrixgame3, swm, alayaworld}. These generators produce a clean latent estimate at every denoising step, not only the last, so an intermediate estimate can serve as a final output.

\paragrapht{Adaptive Denoising.} Caching methods accelerate diffusion by reusing intermediate computation across denoising steps~\citep{deepcache, deltadit, fora, pab, teacache}, including reuse selected per token~\citep{toca} and, in AR video generation, per chunk or by local motion~\citep{flowcache, motioncache}. Spatially adaptive samplers redistribute computation within the schedule, with RAS recomputing only the regions the model attends to~\citep{ras} and HSA evaluating each token at a different number of steps along the same noise schedule~\citep{hsa}. In both cases every token is carried to the end of the schedule, so no token is ever declared finished, and the error left by the computation that was skipped has to be kept bounded by periodic full updates. Substituting a previous output also weakens as the schedule shortens, since distilled samplers leave fewer steps over which one output can stand in for the next~\citep{disca}.

In Self Forcing-style generators, DSA and Surprise Forcing select denoising depth for an entire frame or chunk, assigning the same number of steps to settled and unsettled regions~\citep{dsa, surpriseforcing}. Speculative Forcing instead treats intermediate clean estimates as drafts that can be committed at different steps according to token-wise convergence (Figure~\ref{fig:diff_map}). It draws on draft verification in speculative decoding~\citep{leviathan, chen, selfspec}, without adopting its exact acceptance rule. Committed tokens receive no further denoising updates, while their cached keys and values remain as context for the remaining tokens.

\section{Method}
\label{sec:method}
Speculative Forcing allows individual tokens to stop denoising at different steps. Accepted drafts are fixed as output, while their cached keys and values provide context for tokens that continue denoising (Section~\ref{sec:framework}). To determine when to accept each draft, we introduce two verifiers that predict convergence from latent estimates or the generator's intermediate features (Section~\ref{sec:verification}).

\subsection{Preliminaries}
\label{sec:prelim}

\paragrapht{Autoregressive video diffusion.}
Autoregressive (AR) video diffusion models generate a video as a sequence of latent chunks $\{c_1, c_2, \dots\}$, each containing one or more consecutive frames. The joint distribution is factorized as
$ p(c_1, c_2, \dots)
=
\prod_t p(c_t \mid c_{<t}),$ 
where each conditional distribution is modeled through a diffusion process. At inference, the model generates each chunk by progressively denoising Gaussian noise conditioned on previously generated chunks. We consider generators that use clean history $\mathcal{H}_t = c_{<t}$ as conditioning and cache its keys and values to avoid recomputing history features during subsequent generation.

\paragrapht{Few-step autoregressive generation.}
Distilled AR video generators~\citep{causvid, selfforcing, longlive} reduce the denoising process to $K$ steps, enabling real-time streaming. Each chunk contains $N$ spatiotemporal latent tokens indexed by $i$. At step $k \in \{1,\dots,K\}$, the generator takes the current noisy latent and the clean history as input and predicts a clean estimate $\hat{x}_0^{(k)}$. For $k<K$, this estimate is re-noised to the next noise level for further denoising, and the final estimate $\hat{x}_0^{(K)}$ is returned as the generated chunk.

These generators can be trained with distribution matching distillation~\citep{dmd, dmd2}, which aligns their output distribution with a target distribution represented by a pretrained teacher. In particular, Self Forcing~\citep{selfforcing} trains on autoregressive rollouts conditioned on self-generated history and randomly selects a denoising exit step at which to evaluate the distribution-matching objective. Intermediate clean estimates thus receive supervision as generated samples, although the standard inference procedure runs every token through all $K$ steps.

\subsection{Token-wise Commitment}
\label{sec:framework}

\begin{figure}
\centering
\includegraphics[width=.88\linewidth]{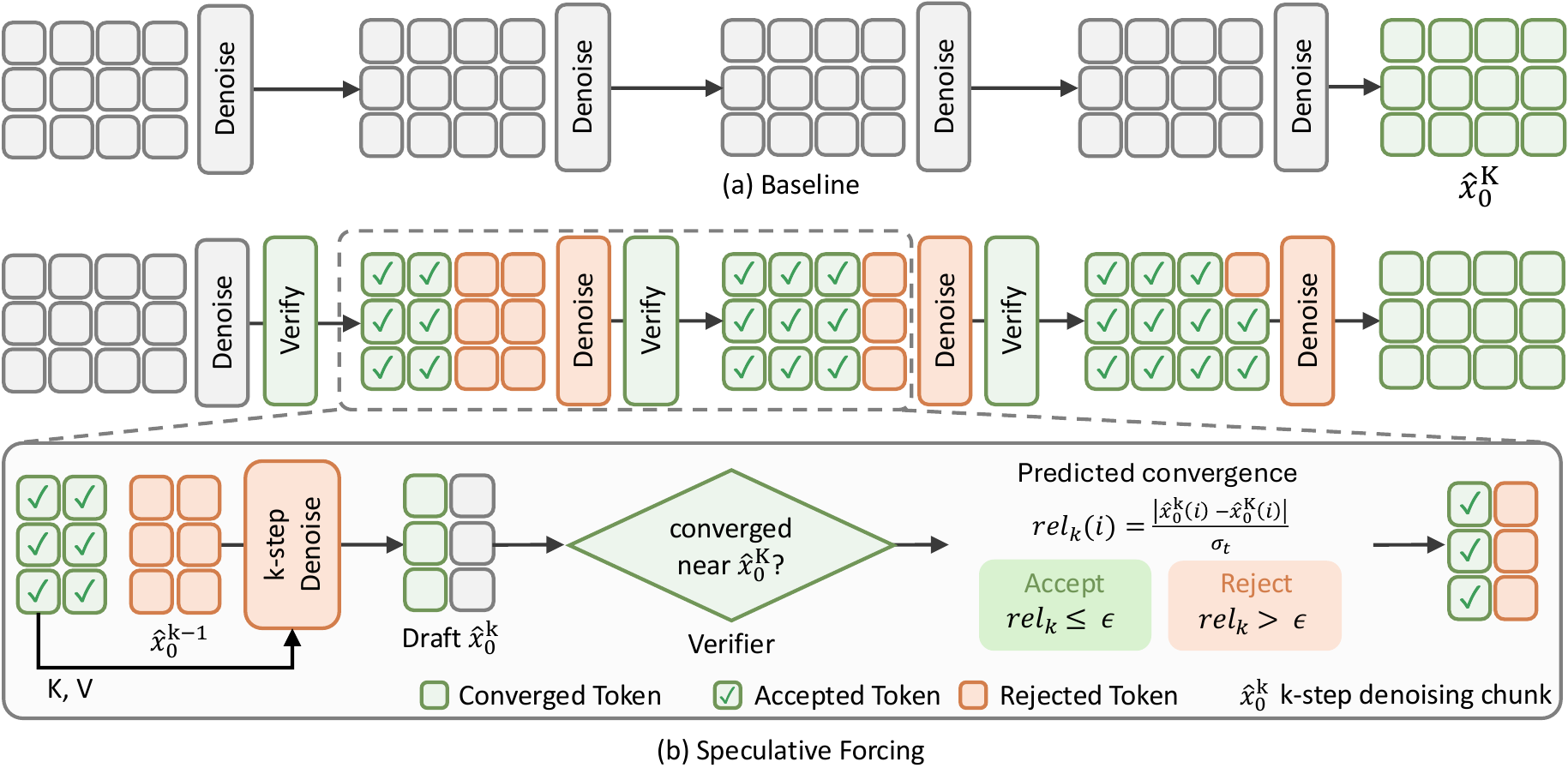}
\vspace{-1em}
\caption{
\textbf{Speculative Forcing.} Baseline runs every token through the full denoising schedule. Speculative Forcing instead treats each token's intermediate clean estimate as a draft and asks a verifier to predict its remaining gap $\mathrm{rel}_k(i)$ to the full-step output. Tokens predicted to have converged are accepted and committed, after which they enter the next step only as keys and values, so the following denoising pass is computed for the rejected tokens alone.
}
\vspace{-1em}
\label{fig:pipeline}
\end{figure}

Figure~\ref{fig:diff_map} shows that tokens approach the full-step output at different rates: some intermediate estimates already closely match their final values, while others continue to change. This motivates committing converged tokens while continuing to refine the rest. 
Speculative Forcing treats each intermediate token estimate as a draft and uses a verifier to predict whether further denoising would substantially change it (Figure~\ref{fig:pipeline}). Accepted drafts are committed as final outputs and removed from further denoising, while the remaining tokens continue to be refined. 

\paragrapht{Commitment and continued denoising.}
Every token starts active, and the first step produces a draft for every token. At each step $k<K$, the verifier evaluates the active tokens. An accepted token $i$ has its output fixed to $\hat{x}_0^{(k)}(i)$ and is excluded from all subsequent denoising forwards. Its layer-wise keys and values computed at commitment remain cached, allowing active tokens to continue attending to it. Rejected tokens are re-noised to the next noise level and continue denoising. Any tokens still active at step $K$ are committed at that step.

Let $k^\ast(i)$ denote the commit step of token $i$. The completed chunk combines estimates accepted at potentially different steps,
$ c_t(i)
=
\hat{x}_0^{(k^\ast(i))}(i).$
The base generator's clean pass over $c_t$ then refreshes the KV cache for subsequent chunks. Commitment reduces the number of denoising token evaluations from $KN$ to $\sum_{i=1}^{N}k^\ast(i)$, with mean denoising depth $\bar{k}=\frac{1}{N}\sum_{i=1}^{N}k^\ast(i)$. Both methods additionally perform $N$ token evaluations for the clean pass.

\paragrapht{Convergence criterion.}
To decide which drafts can be accepted, we measure how closely an intermediate estimate agrees with the generator's full-step output. Here, $\hat{x}_0^{(K)}$ denotes the output of the original generator with every token evaluated at all $K$ steps, using the same initial noise and conditioning. This reference measures the change that remains under the original denoising schedule.

We normalize the distance by $\sigma_t$, the median token norm in the full-step output of chunk $t$, and define the convergence gap as
\begin{equation}
\mathrm{rel}_k(i)
=
\frac{
\left\lVert
\hat{x}_0^{(k)}(i)-\hat{x}_0^{(K)}(i)
\right\rVert_2
}{
\sigma_t
}, 
\qquad \ \ \text{for} \ \ k<K.
\label{eq:rel}
\end{equation}
A draft meets the convergence criterion when $\mathrm{rel}_k(i)\le\epsilon$. Since the full-step output is unavailable during early commitment, this gap serves as an oracle and is used as the ground truth in \cref{eq:head_loss}.

\subsection{Convergence Verification}
\label{sec:verification}
To make acceptance decisions before the full-step output is available, we seek evidence of convergence in the current draft and the features used to produce it. The draft can be compared with clean content or its own previous estimate, motivating a training-free verifier. The generator's intermediate features provide additional contextual information, which a learned verifier uses to predict the convergence gap. 
Since latent tokens encode content across multiple output frames, isolated acceptance decisions can affect both spatial and temporal coherence. We therefore apply a spatiotemporal smoothing operator $\mathcal{S}$ to each verifier's scores before thresholding, encouraging coherent decisions across neighboring tokens.

\subsubsection{Training-Free Verifier}
\label{sec:tf}

At the first step, only one draft exists, so a token's own change cannot be observed. We therefore rely on its redundancy with content that is already clean. From the second step on, we judge convergence by how much the token's consecutive drafts differ.

\paragrapht{Redundancy-based verification.}
We observe that drafts resembling previously generated or reference content often have smaller remaining changes (Figure~\ref{fig:diff_map}). This motivates similarity to clean content as a proxy for convergence. At the first step, we score each token by the distance from its draft to the most similar clean latent token, $
s_{\mathrm{red}}(i)
=
\min_{h \in \mathcal{H}}
\left\lVert
\hat{x}_0^{(1)}(i)-h
\right\rVert_2,
$ where $\mathcal{H}$ contains clean latent tokens from previous chunks within the generator's attention window and any reference frames, including those retained as attention sinks. The search covers all spatial positions within this bank to accommodate camera and object motion. Tokens with $\mathcal{S}[s_{\mathrm{red}}](i)\le\tau_{\mathrm{red}}$ are accepted.

\paragrapht{Stability-based verification.}
From the second step on, two consecutive drafts of the same active token are available. Their difference measures how much the token's estimate has changed over the latest denoising step, $
s_{\mathrm{stab}}^{(k)}(i)
=
\left\lVert
\hat{x}_0^{(k)}(i)-\hat{x}_0^{(k-1)}(i)
\right\rVert_2, \ \ \text{for} \ \ 2 \le k < K.
$ Active tokens with $\mathcal{S}[s_{\mathrm{stab}}^{(k)}](i)\le\tau_{\mathrm{stab}}$ are accepted. Both thresholds are calibrated on held-out data. Tokens whose drafts closely match clean content can thus be committed at the first step, while tokens with sufficiently small changes between consecutive drafts can be committed at later steps. Tokens that satisfy neither condition run the full schedule.

\subsubsection{Learned Verifier}
\label{sec:head}

Redundancy and stability are proxies for convergence rather than convergence itself. A token can match the history and still need refinement, or change little between two steps and yet remain far from its final value. The learned verifier instead predicts the convergence gap directly from the generator's own features. A shallow convolutional head suffices for this prediction, indicating that the intermediate features contain information about each token's remaining discrepancy to the full-step output. 

\paragrapht{Training.}
The head takes the hidden state of a selected backbone layer at step $k$ and outputs a predicted gap $\hat{r}_k(i)$ for each token. It consists of layer normalization followed by three 3D convolutions and is shared across all steps. We run the frozen generator for the full $K$ steps and collect intermediate hidden states together with their oracle convergence gaps. Only the head is trained, using a Huber loss,
\begin{equation}
\mathcal{L}(\phi)
=
\frac{1}{N(K-1)}
\sum_{i=1}^{N}
\sum_{k=1}^{K-1}
\mathrm{smooth}_{\ell_1}
\bigl(
\hat{r}_k(i),\,\mathrm{rel}_k(i)
\bigr),
\label{eq:head_loss}
\end{equation}
where $\phi$ denotes the head parameters.

\paragrapht{Verification.}
At inference, we retain the last hidden states of committed tokens and update those of active tokens, preserving the token grid for the convolutional head. At each step $k<K$, the head predicts $\hat{r}_k(i)$, and active tokens with $\mathcal{S}[\hat{r}_k](i)\le\epsilon$ are accepted.

Because the head predicts the convergence gap itself, $\epsilon$ sets a tolerance on the predicted discrepancy to the full-step output. Local smoothing encourages spatially and temporally coherent decisions. Adjusting $\epsilon$ controls the speed-quality trade-off at inference without retraining or fine-tuning.

\section{Experiment}

\label{sec:experiments}

We evaluate Speculative Forcing on autoregressive video generation and world modeling. We compare generation quality and inference speed against uniform-step sampling, and examine whether preserving settled predictions while selectively refining the remaining tokens reduces error drift during autoregressive rollouts. We further analyze the verification tolerance, token selection, and the granularity of step allocation. 

\subsection{Experimental Details}
\label{sec:exp_details}

\paragrapht{Experimental setups.} 
We apply our method to LongLive~\citep{longlive} (1.3B) and the Causal-Fast model of Lingbot-World 2.0~\citep{lingbotworld2} (1.3B), both using four denoising steps. We keep both generators frozen and evaluate training-free and learned verifiers. For LongLive, we train the learned verifier on full-step rollouts from 1K VidProM~\citep{vidprom} prompts following the generator's original training rollout procedure. For Lingbot, we collect training rollouts from 1K initial images with scenes disjoint from evaluation, using the same random camera-trajectory sampling and rollout protocol as evaluation. Both use the same 3-layer 3D convolutional head with negligible overhead.urs$_\mathrm{H}$ uses the head verifier, while Ours$_\mathrm{TF}$ uses the training-free verifier.

The verification tolerance can be adjusted at inference without retraining. We set the learned verifier's tolerance to $\epsilon=0.28$ for short videos and use a more conservative $\epsilon=0.24$ for long videos and world model rollouts. For each setting, we calibrate the training-free thresholds $(\tau_{\mathrm{red}},\tau_{\mathrm{stab}})$ on held-out calibration data against the oracle convergence criterion $\mathrm{rel}_k(i)\le\epsilon$, using the same tolerance as the learned verifier.
We compare against uniform sampling with one, two, three, and four denoising steps and report mean denoising steps and throughput on a single NVIDIA H100. Implementation details are provided in Appendix~\ref{app:implementation}.

\paragrapht{Evaluation.}
For LongLive, we evaluate 5-second videos using the VBench prompts rewritten by Self Forcing~\citep{selfforcing}, with five random seeds per prompt, and report all 16 VBench dimensions~\citep{vbench}. We also generate 30-second videos with the same prompt set and evaluate all 16 dimensions using VBench-Long~\citep{vbenchpp}, following the protocol of LongLive~\citep{longlive}.

We further evaluate Lingbot on 200 held-out navigation scenes to examine performance under substantial changes in visible content and pronounced intermediate distortions. Each scene provides an initial image, a text prompt, and a camera trajectory composed of five segments with randomly sampled camera actions and motion speeds. We generate 253 frames per scene and report a VBench quality aggregate over six dimensions, excluding temporal flickering. We also report relative camera-pose error against the commanded ground-truth trajectory to assess whether accelerated sampling follows the camera path as faithfully as the four-step generator. For both long videos and world model rollouts, we additionally report the Drift metric of Reward Forcing~\citep{rewardforcing}, which measures temporal variation in imaging quality across video segments. All methods share the same conditioning inputs and sampling seeds. Metric definitions and benchmark construction are detailed in Appendix~\ref{app:evaluation}.

\subsection{Experimental Results}
\label{sec:results}

\begin{wraptable}{r}{0.46\textwidth}
\centering
\footnotesize
\tabcolsep=0.2em
\renewcommand{\arraystretch}{1.2}
\caption{\textbf{Comparison on 5s videos with LongLive.} Mean Steps is the measured average number of denoising passes per token. Standard VBench 16-dimension protocol, throughput on a single H100.}
\label{tab:main_5s}
\vspace{-1em}
\begin{tabular}{l c c c c c}
\toprule
\multirow{2}{*}{Method} &
\multirow{2}{*}{\makecell{Mean\\Steps $\downarrow$}} &
\multirow{2}{*}{\makecell{Through.\\(FPS)}} &
\multicolumn{3}{c}{Evaluation scores $\uparrow$}\\
\cmidrule(lr){4-6}
 & & & Total & Quality & Semantic \\
\midrule
\rowcolor{gray!12}\multicolumn{6}{l}{\textit{LongLive~\citep{longlive}}}\\
1-step                          & 1.00 & 67.97 & 79.06 & 80.03 & 75.17 \\
2-step                          & 2.00 & 44.54 & 82.46 & 83.06 & 80.09 \\
3-step                          & 3.00 & 33.15 & 83.08 & 83.45 & \underline{81.58} \\
4-step                          & 4.00 & 26.40 & 83.08 & 83.49 & 81.43 \\
\rowcolor{myred}
\textbf{Ours\textsubscript{TF}} & 1.65 & 43.61 & \underline{83.12} & \underline{83.51} & 81.56 \\
\rowcolor{myblue}
\textbf{Ours\textsubscript{H}}  & 1.68 & 41.74 & \textbf{83.21} & \textbf{83.54} & \textbf{81.89} \\
\bottomrule
\end{tabular}
\vspace{-1.5em} 
\end{wraptable}

\paragrapht{Short video generation.}
Table~\ref{tab:main_5s} compares generation quality and efficiency on 5-second videos. Both verifiers maintain the overall VBench performance of the four-step generator while using fewer than 1.7 denoising steps per token on average. By accepting settled predictions early and continuing to refine the remaining tokens, our method preserves generation quality with less than half the denoising token evaluations of the original schedule.

\begin{table*}[t]
\centering
\scriptsize
\tabcolsep=0.3em
\renewcommand{\arraystretch}{1.15}
\caption{\textbf{Comparison on 30s long videos with LongLive.} Evaluated on VBench-Long.}
\label{tab:main_30s}
\vspace{-1em} 
\begin{tabular}{l c c c c c c c c c c c c}
\toprule
\multirow{2}{*}{Method} &
\multirow{2}{*}{\makecell{Mean\\Steps $\downarrow$}} &
\multirow{2}{*}{\makecell{Through.\\(FPS)}} &
\multirow{2}{*}{\makecell{Drift\\$\downarrow$}} &
\multirow{2}{*}{\makecell{Total\\Score $\uparrow$}} &
\multirow{2}{*}{\makecell{Quality\\Score $\uparrow$}} &
\multirow{2}{*}{\makecell{Semantic\\Score $\uparrow$}} &
\multicolumn{6}{c}{VBench-Long dimension scores $\uparrow$}\\
\cmidrule(lr){8-13}
 & & & & & & &
\makecell{Subject\\Consistency} & \makecell{Background\\Consistency} &
\makecell{Motion\\Smoothness} & \makecell{Dynamic\\Degree} &
\makecell{Aesthetic\\Quality} & \makecell{Imaging\\Quality}\\
\midrule
\rowcolor{gray!12}\multicolumn{13}{l}{\textit{LongLive~\citep{longlive}}}\\
1-step                          & 1.00 & 61.6 & 2.930 & 80.82 & 81.87 & 76.64 & 96.94 & 95.76 & 98.24 & \textbf{49.63} & 60.09 & 63.81 \\
2-step                          & 2.00 & 40.9 & 1.992 & 82.32 & 82.99 & 79.61 & 97.67 & 96.38 & 98.60 & \underline{44.81} & 62.92 & 67.66 \\
3-step                          & 3.00 & 30.6 & 1.738 & 82.60 & 83.09 & \underline{80.67} & 98.09 & 96.75 & \underline{98.77} & 38.98 & 63.71 & \textbf{68.69} \\
4-step                          & 4.00 & 24.5 & 1.722 & 82.69 & \underline{83.29} & 80.30 & \textbf{98.22} & \textbf{96.99} & \textbf{98.85} & 39.17 & \textbf{64.43} & 68.52 \\
\rowcolor{myred}
\textbf{Ours\textsubscript{TF}} & 1.71 & 38.8 & \textbf{1.685} & \underline{82.75} & 83.28 & 80.65 & \underline{98.11} & 96.77 & 98.74 & 41.57 & 63.83 & 68.27 \\
\rowcolor{myblue}
\textbf{Ours\textsubscript{H}}  & 1.72 & 37.8 & \underline{1.707} & \textbf{82.81} & \textbf{83.32} & \textbf{80.77} & \textbf{98.22} & \underline{96.79} & 98.75 & 40.74 & \underline{63.95} & \underline{68.66} \\
\bottomrule
\end{tabular}
\vspace{-1.25em} 
\end{table*}

\paragrapht{Long video generation.}
Table~\ref{tab:main_30s} reports results on long video generation. The training-free verifier retains competitive quality with a small loss in consistency, while the learned verifier maintains both the quality and consistency of the four-step generator. Both verifiers use approximately 1.7 denoising steps per token on average and achieve lower Drift than the four-step generator. Both also maintain comparable dynamic degree, suggesting that the reduced quality variation is not simply a consequence of suppressing motion.

\begin{table*}[t]
\centering
\scriptsize
\tabcolsep=0.35em
\renewcommand{\arraystretch}{1.15}
\caption{\textbf{World-model comparison on dynamic videos.} Results with the autoregressive Lingbot-World 2.0~\citep{lingbotworld2}, evaluated on VBench. RPE is the relative pose error of the camera trajectory. Quality Score aggregates the six dimensions reported here.}
\label{tab:main_lingbot}
\vspace{-1em}
\begin{tabular}{l c c c c c c c c c c c c}
\toprule
\multirow{2}{*}{Method} &
\multirow{2}{*}{\makecell{Mean\\Steps $\downarrow$}} &
\multirow{2}{*}{\makecell{Through.\\(FPS)}} &
\multicolumn{2}{c}{RPE} &
\multirow{2}{*}{\makecell{Drift\\$\downarrow$}} &
\multirow{2}{*}{\makecell{Quality\\Score $\uparrow$}} &
\multicolumn{6}{c}{VBench dimension scores $\uparrow$}\\
\cmidrule(lr){4-5}\cmidrule(lr){8-13}
 & & & Trans $\downarrow$ & Rot $\downarrow$ & & &
\makecell{Subject\\Consistency} & \makecell{Background\\Consistency} &
\makecell{Motion\\Smoothness} & \makecell{Dynamic\\Degree} &
\makecell{Aesthetic\\Quality} & \makecell{Imaging\\Quality}\\
\midrule
\rowcolor{gray!12}\multicolumn{13}{l}{\textit{Lingbot-World 2.0~\citep{lingbotworld2}}}\\
1-step                          & 1.00 & 46.9 & 0.552 & 0.360 & 5.579 & 77.19 & 81.31 & 87.23 & 96.98 & 96.50 & 58.38 & 66.63 \\
2-step                          & 2.00 & 30.7 & 0.345 & 0.255 & 3.999 & 81.68 & 87.71 & 90.28 & 97.10 & 98.50 & 65.98 & 70.65 \\
3-step                          & 3.00 & 22.8 & \underline{0.329} & \textbf{0.251} & 3.361 & 82.38 & 88.42 & \underline{90.67} & 97.13 & \textbf{99.00} & 67.14 & 71.61 \\
4-step                          & 4.00 & 18.1 & \textbf{0.328} & \underline{0.253} & 3.094 & \underline{82.62} & \underline{88.50} & \textbf{90.83} & \textbf{97.23} & \textbf{99.00} & \underline{67.18} & 72.24 \\
\rowcolor{myred}
\textbf{Ours\textsubscript{TF}} & 2.53 & 24.7 & \underline{0.329} & 0.259 & \underline{2.926} & 82.58 & 88.47 & 90.57 & \underline{97.16} & \textbf{99.00} & 67.17 & \textbf{72.69} \\
\rowcolor{myblue}
\textbf{Ours\textsubscript{H}}  & 2.55 & 24.5 & \textbf{0.328} & 0.260 & \textbf{2.839} & \textbf{82.65} & \textbf{88.56} & 90.62 & 97.12 & \textbf{99.00} & \textbf{67.55} & \underline{72.65} \\
\bottomrule
\end{tabular}
\vspace{-1.5em}
\end{table*}

\begin{figure}[t!]
\centering
\includegraphics[width=.95\linewidth]{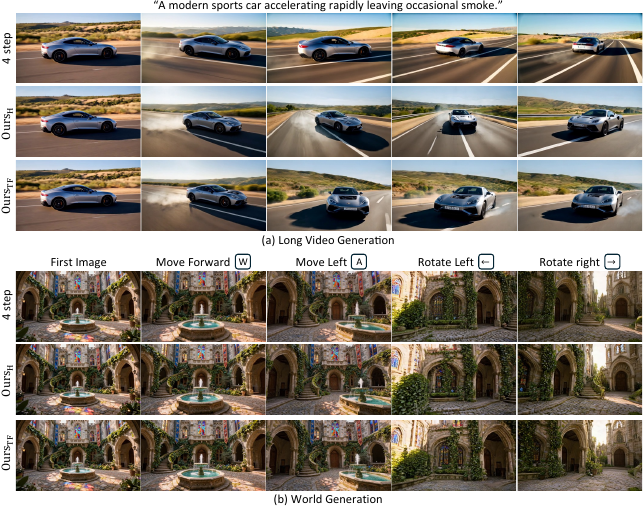}
\vspace{-.5em}
\caption{
\textbf{Qualitative comparisons on video generation and world modeling.}
Both our head and training-free verifiers preserve visual quality comparable to the four-step baseline without visibly increasing quality drift over the illustrated rollouts on our benchmark.
}
\vspace{-1em}
\label{fig:main_qual}
\end{figure}

\paragrapht{World generation.}
Table~\ref{tab:main_lingbot} reports results on world generation, a more challenging setting in which camera control induces substantial changes across the visible scene. Accordingly, our methods tend to allocate more denoising steps than in standard video generation. Despite this increased difficulty, the training-free verifier remains competitive in quality while maintaining low Drift. The learned verifier preserves quality and consistency comparable to the four-step generator while achieving lower Drift. Camera-pose errors against the commanded trajectory also remain comparable, indicating that acceleration preserves camera-following accuracy.

\paragrapht{Qualitative results.}
Figure~\ref{fig:main_qual} compares our generations with the corresponding four-step generators. On both LongLive and Lingbot, our method preserves visual quality comparable to four-step generation without noticeable degradation over time. On LongLive, it maintains subject identity and image quality even in scenes with dynamic objects. On Lingbot, previously observed regions remain coherent under large camera motion, while newly revealed regions are further refined, yielding consistent transitions between accepted and refined areas.

\subsection{Ablation Studies}
\label{sec:ablation}

\begin{figure}
\centering
\includegraphics[width=.95\linewidth]{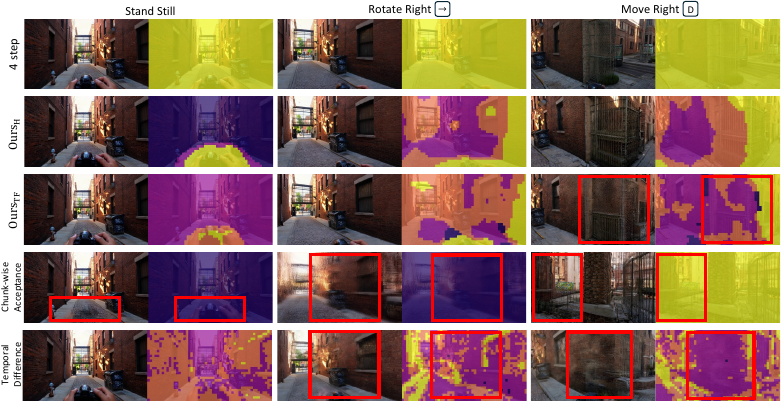}
\vspace{-.5em}
\caption{
\textbf{Qualitative ablations of verification and commitment granularity.}
Our learned verifier handles distortions that challenge training-free verification.
Premature chunk-wise acceptance propagates errors to later chunks, while motion-based verification struggles to distinguish camera-induced changes from distortions.
}
\vspace{-.5em}
\label{fig:abl_qual}
\end{figure}

We examine how verification tolerance, verification signals, and acceptance granularity affect generation quality and efficiency. Figure~\ref{fig:abl_qual} provides representative qualitative comparisons of the alternative policies. We further evaluate how accurately each verification signal predicts the oracle in Appendix~\ref{app:analysis_verification} and provide additional ablations on our head in Appendix~\ref{app:additional_ablation}.

\paragrapht{Effect of verification tolerance.}
Table~\ref{tab:abl_eps} shows how $\epsilon$ controls the quality-efficiency trade-off on Lingbot. Increasing $\epsilon$ reduces mean denoising steps, with $\epsilon=0.20$ and $\epsilon=0.24$ maintaining the four-step generator's quality while achieving lower Drift. Larger tolerances accept insufficiently refined drafts and degrade quality, highlighting the need to balance early acceptance with further refinement.

\paragrapht{Effect of verification signals.}
We adapt two existing selection criteria to token-wise commitment in latent space. Inspired by MotionCache~\citep{motioncache}, we measure temporal differences between adjacent latent frames at the same spatial position. We also adapt RAS's predicted-noise dispersion signal~\citep{ras} to the channel variance of each predicted clean latent token. Thresholds are calibrated on held-out data to obtain comparable mean denoising steps. As shown in Table~\ref{tab:abl_comp}, both adapted signals yield lower quality than our learned verifier and uniform sampling at similar mean steps. Temporal differences, for example, can reflect both camera-induced changes and intermediate distortions, limiting their reliability as convergence signals (Figure~\ref{fig:abl_qual}). These results support directly predicting each draft's remaining convergence gap.

\paragrapht{Comparison with oracle verification.}
We additionally evaluate an oracle verifier that computes the full-step reference for each chunk using the same initial noise and conditioning, then uses the measured convergence gaps to determine token acceptance. We apply the same tolerance and spatiotemporal smoothing as the learned verifier. Our learned verifier achieves generation quality comparable to oracle verification without computing the full-step reference at inference (Table~\ref{tab:abl_comp}).

\begin{table}[t]
\centering
\begin{minipage}[t]{0.49\textwidth}

\centering
\caption{One test-time threshold sweeps speed against quality. At
$\epsilon \in [0.20, 0.24]$ our sampler exceeds the uniform 4-step baseline on quality metrics. Drift is lowest in this range, not at the full budget.} 
\label{tab:abl_eps}
\vspace{-1em}
\tabcolsep=0.1em
\renewcommand{\arraystretch}{1.2}
\resizebox{\linewidth}{!}{%
\begin{tabular}{l c c c c c}
\toprule
Tolerance $\epsilon$ & \makecell{Mean\\Steps $\downarrow$} & \makecell{Through.\\(FPS) $\uparrow$} & \makecell{Quality $\uparrow$} & \makecell{Imaging\\Quality $\uparrow$} & \makecell{Drift\\$\downarrow$}\\
\midrule
$\epsilon = 0.16$ & 3.47 & 20.1 & 82.56 & 72.43 & 3.010 \\
$\epsilon = 0.20$ & 2.95 & 22.4 & \textbf{82.72} & \underline{72.64} & \underline{2.862} \\
$\epsilon = 0.24$ & 2.55 & 24.5 & \underline{82.65} & \textbf{72.65} & \textbf{2.839} \\
$\epsilon = 0.28$ & 2.24 & 26.4 & 82.53 & 72.32 & 2.943 \\
$\epsilon = 0.32$ & 1.97 & 28.3 & 82.07 & 71.69 & 3.398 \\
\midrule
\textit{Uniform 4-step} & 4.00 & 18.1 & 82.62 & 72.24 & 3.094 \\
\bottomrule
\end{tabular}}

\end{minipage}\hfill
\begin{minipage}[t]{0.49\textwidth}

\centering
\caption{Routing-strategy comparison on world generation benchmark on Lingbot-World 2.0. Every method is recalibrated on $\epsilon = 0.24$. \textit{Oracle} measures the true residual by running all steps for each chunk; therefore, we report no throughput (N/A).}
\label{tab:abl_comp}
\vspace{-1em}
\tabcolsep=0.1em
\renewcommand{\arraystretch}{1.2}
\resizebox{\linewidth}{!}{%
\begin{tabular}{l c c c c c}
\toprule
Method & \makecell{Mean\\Steps $\downarrow$} & \makecell{Through.\\(FPS) $\uparrow$} & \makecell{Quality $\uparrow$} & \makecell{Imaging\\Quality $\uparrow$} & \makecell{Drift\\$\downarrow$}\\
\midrule
Channel Variance & 2.60 & 24.2 & 78.23 & 66.59 & 4.996 \\
Temporal Difference & 2.58 & 24.4 & 80.97 & 70.22 & 4.116 \\
Chunk-wise Acceptance & 2.61 & 25.2 & 82.54 & 72.26 & 3.214 \\
\rowcolor{myred}
\textbf{Ours\textsubscript{TF}} & 2.53 & 24.7 & \underline{82.58} & \textbf{72.69} & \underline{2.926} \\
\rowcolor{myblue}
\textbf{Ours\textsubscript{H}} & 2.55 & 24.5 & \textbf{82.65} & \underline{72.65} & \textbf{2.839} \\
\midrule
\textit{Oracle} & 2.46 & (N/A) & 82.61 & 72.60 & 2.903 \\
\textit{Uniform 4-step} & 4.00 & 18.1 & 82.62 & 72.24 & 3.094 \\
\bottomrule
\end{tabular}}

\end{minipage}
\vspace{-1em}
\end{table}

\paragrapht{Effect of token-wise commitment.}
Adaptive denoising has been explored at frame and chunk granularity~\citep{dsa, surpriseforcing}. To isolate commitment granularity, we use the same learned verifier but accept an entire chunk when its mean predicted convergence gap falls below a threshold at each step. The threshold is calibrated on held-out data against the oracle criterion to match our mean denoising steps. As shown in Table~\ref{tab:abl_comp}, token-wise commitment achieves higher quality and lower Drift at comparable mean denoising steps. Chunk-wise acceptance cannot distinguish tokens that have already converged from those that continue to change, resulting in lower quality and higher Drift when all tokens share the same acceptance step (Figure~\ref{fig:abl_qual}).

\section{Conclusion}
\label{sec:conclusion}

We introduced \textbf{Speculative Forcing}, a framework that allocates denoising computation according to token-wise convergence in few-step autoregressive video generation. By treating intermediate clean estimates as drafts, it predicts which tokens can be accepted early and continues denoising only the remaining tokens. Training-free and learned verifiers enable these decisions without modifying the generator's weights. Experiments on short and long videos and controllable world model rollouts demonstrate faster generation while maintaining quality, with comparable or lower quality variation and preserved camera-following accuracy in world modeling. These results show that a shared denoising depth is not necessary for every token, supporting convergence-based acceptance as a practical approach to efficient video generation.

\bibliography{iclr2027_conference}
\bibliographystyle{iclr2027_conference}
\appendix
\newpage
\renewcommand\thefigure{\Alph{figure}}    
\setcounter{figure}{0}  
\renewcommand\thetable{\Alph{table}}
\setcounter{table}{0} 
\section*{\Large Appendix}
{
    \renewcommand{\baselinestretch}{1.8}\selectfont
    
    \noindent
    \textbf{A\quad Implementation Details}

    \paragrapht{B\quad Evaluation Details}
    
    \hspace*{2em}B.1\quad Evaluation Metrics 
    \vspace{-5pt}

    \hspace*{2em}B.2\quad Benchmark Construction 
    \vspace{-5pt}   
    
    \paragrapht{C\quad Additional Results and Analyses}
    
    \hspace*{2em}C.1\quad Analysis of Verification Accuracy 
    
    \vspace{-5pt}

    \hspace*{2em}C.2\quad Additional Ablation on Our Head
    
    \vspace{-5pt}

    \hspace*{2em}C.3\quad Additional Quantitative Results 
    \vspace{-5pt}
    
    \hspace*{2em}C.4\quad Additional Qualitative Results 

    \vspace{-5pt}
    
    \paragrapht{D\quad Discussions}
    
    \hspace*{2em}D.1\quad Limitations
    \vspace{-5pt}
    
    \hspace*{2em}D.2\quad Other Discussions 

    \vspace{-5pt}
    
    \hspace*{2em}D.3\quad Potential Applications

    \vspace{-5pt}
    
    \vspace{10pt}

}

\vfill

\clearpage

\section{Implementation Details}
\label{app:implementation}

\label{app:baseline_adaptation}
We retain the released backbone configurations except for the denoising policies evaluated in this work. For LongLive~\citep{longlive}, we use an attention sink of 3 latent frames, a local attention window of 12 latent frames, and 3 latent frames per chunk. Short and long videos contain 81 and 477 frames, respectively. For Lingbot-World 2.0~\citep{lingbotworld2}, we use the 1.3B Causal-Fast model with an attention sink of 6 latent frames, a local attention window of 18 latent frames, and 4 latent frames per chunk. Each benchmark video contains 253 frames. Lingbot-World 2.0 additionally receives a per-frame camera trajectory encoded through Pl\"ucker embeddings derived from relative camera-to-world poses. We retain the original camera-conditioning pathway, including per-video translation normalization by the maximum translation magnitude. For the main results, the training-free thresholds $(\tau_{\mathrm{red}},\tau_{\mathrm{stab}})$ are $(0.497,0.327)$ for 5-second LongLive videos, $(0.357,0.235)$ for 30-second LongLive videos, and $(0.375,0.300)$ for Lingbot-World 2.0.

We retain the original denoising schedules so that early commitment operates on the intermediate clean estimates supervised during distribution-matching training. In our configurations, LongLive and Lingbot-World 2.0 share the shifted timesteps $[1000.0,937.5,833.3,625.0]$, obtained from a 1000-step flow-matching schedule with a shift of $5.0$. A token accepted at step $k$ undergoes exactly the first $k$ steps of this schedule. Uniform-step baselines use the same prefixes, allowing us to compare token-wise acceptance with a shared exit step along the same trained denoising trajectory.

The learned verifier reads hidden states from the generator's final transformer block without an additional backbone forward. The head consists of layer normalization and three 3D convolutions with channel dimensions $1536\to96\to48\to1$. The first two convolutions use $3\times3\times3$ kernels with GELU activations, and the last uses a $1\times1\times1$ kernel. The head's output is average-pooled over $2\times2\times2$ token blocks and expanded back to the original grid using nearest-neighbor interpolation. The head has 4.11M parameters and is shared across denoising steps. We train only the head using the smooth-$\ell_1$ loss and Adam with a learning rate of $10^{-3}$ and batch size 4. 

Within each chunk, accepted tokens retain the layer-wise keys and values computed at their acceptance step, while active tokens update theirs as denoising continues. The cache can therefore contain KV states from different denoising steps. Once the chunk is complete, the generator performs a clean forward pass over the assembled output at timestep $0$, refreshing the KV states of all tokens regardless of their acceptance steps. Subsequent chunks thus attend to history states computed from the completed clean output.

We apply spatiotemporal smoothing to each verifier's score map before thresholding at every verification step. Spatial smoothing uses a normalized $9\times9$ Gaussian kernel with a standard deviation of 1.0 token and reflect padding, applied independently to each latent frame. Temporal smoothing then averages scores over the current and subsequent latent frames within the same chunk, using Gaussian weights with a standard deviation of 1.0 latent frame. 

Only tokens that have not yet been accepted are forwarded through the generator at subsequent denoising steps. We reduce execution overhead through shared indexing and in-place KV updates. We also merge each chunk's clean pass with the next chunk's first denoising pass while preserving timestep conditioning, attention dependencies, and KV update order. All baselines use the same applicable execution optimizations for fair throughput comparisons.

\section{Evaluation Details}
\label{app:evaluation}

\subsection{Evaluation Metrics}
\label{app:evaluation_metrics}
We follow VBench~\citep{vbench} for 5-second videos and VBench-Long~\citep{vbenchpp} for 30-second videos for LongLive, reporting all 16 dimensions. For Lingbot-World 2.0, we aggregate six quality dimensions using VBench's official normalization and weights, excluding temporal flickering because intended camera motion can produce frame-to-frame pixel changes. Following Reward Forcing~\citep{rewardforcing}, we measure Drift as the standard deviation of imaging-quality scores across non-overlapping 2-second segments, averaged over videos. We use VBench's MUSIQ model and preprocessing, with 7 complete segments per Lingbot evaluation and 15 segments per 30-second LongLive evaluation. Lower Drift indicates more consistent imaging quality over time.

For Lingbot-World 2.0, we measure relative pose error (RPE) against the commanded ground-truth camera trajectory. We recover camera poses using Depth Anything 3~\citep{da3} and apply Sim3 Umeyama alignment to the commanded trajectory independently for each of the five camera-control segments, yielding 1,000 segments across 200 videos. We compare relative transformations between consecutive frames. Rot is the mean angular error over all frame pairs, in degrees. Trans is the sum of translation errors divided by the sum of commanded displacements over all pairs, and is dimensionless. This protocol evaluates local camera-following accuracy under the same reconstruction and alignment procedure for all methods.

To quantify denoising computation, we report the average number of denoising evaluations per token across all chunks and videos, excluding the clean pass used to refresh the KV cache. We measure throughput on a single NVIDIA H100 as the number of generated frames divided by the DiT forward time, excluding VAE encoding and decoding.

\subsection{Benchmark Construction}
\label{app:benchmark_construction}

We construct a fixed benchmark of 200 scenes for Lingbot-World 2.0, each consisting of an initial image, a text prompt, and a camera trajectory. All methods use the same conditioning inputs and sampling seeds. Qwen2.5-7B-Instruct~\citep{qwen2.5} generates scene descriptions spanning diverse categories and visual styles, and Z-Image-Turbo~\citep{zimage} generates the initial images at $1376\times768$. The image and video models receive the same scene prompt, combining environment, subject, and perspective descriptions. After fixing the benchmark, we construct the verifier's training set from separate scenes using the same pipeline, filtering potential overlap through text and image similarity.
Each video contains 253 frames, with the 252 camera transitions divided into five consecutive control segments of lengths 50, 50, 50, 50, and 52. For each segment, we uniformly sample an action from forward, backward, strafe left, strafe right, yaw left, yaw right, and stop, excluding the preceding segment's action. Across the benchmark, this produces 593 translation, 276 rotation, and 131 stationary segments. Translation speeds are sampled from $\mathcal{U}(0.2,1.0)$, with the fastest translation segment in each video set to $1.0$ to match the model's per-video translation normalization. Yaw rates are sampled from $\mathcal{U}(0.6,2.0)$ degrees per frame. We choose the lower bound based on preliminary tests in which smaller commands produced little visible rotation. Camera intrinsics follow the official adapter defaults and remain fixed across methods.

\section{Additional Results and Analyses}
\label{app:additional_results}

\subsection{Analysis of Verification Accuracy}
\label{app:analysis_verification}
\begin{table}[t]
\centering
\footnotesize
\setlength{\tabcolsep}{0.4em}
\renewcommand{\arraystretch}{1.2}
\caption{\textbf{Verification accuracy on LongLive.} We report AUROC $\uparrow$ for identifying tokens requiring further refinement, defined by $\mathrm{rel}_k(i)>\epsilon$.}
\label{tab:sup_accuracy}
\vspace{-0.5em}
\resizebox{\linewidth}{!}{%
\begin{tabular}{l ccc ccc ccc}
\toprule
\multirow{2}{*}{Verifier signal}
& \multicolumn{3}{c}{$\epsilon=0.16$}
& \multicolumn{3}{c}{$\epsilon=0.24$}
& \multicolumn{3}{c}{$\epsilon=0.32$}\\
\cmidrule(lr){2-4}
\cmidrule(lr){5-7}
\cmidrule(lr){8-10}
& $k=1$ & $k=2$ & $k=3$
& $k=1$ & $k=2$ & $k=3$
& $k=1$ & $k=2$ & $k=3$\\
\midrule
\rowcolor{gray!12}
\multicolumn{10}{l}{\textit{Learned verifier (head) design}}\\
On-policy training
& 0.932 & 0.909 & 0.905
& 0.935 & 0.889 & 0.902
& 0.941 & 0.871 & 0.887\\
Per-token MLP
& 0.936 & 0.910 & 0.913
& 0.940 & 0.894 & 0.911
& 0.946 & 0.876 & 0.888\\
Ours\textsubscript{H}
& 0.933 & 0.911 & 0.910
& 0.934 & 0.890 & 0.906
& 0.944 & 0.871 & 0.887\\
\midrule
\rowcolor{gray!12}
\multicolumn{10}{l}{\textit{Latent-based verification signals}}\\
Stability
& -- & 0.852 & 0.890
& -- & 0.774 & 0.854
& -- & 0.711 & 0.829\\
Redundancy
& 0.820 & 0.724 & 0.670
& 0.828 & 0.692 & 0.611
& 0.849 & 0.653 & 0.609\\
Motion-based~\citep{motioncache}
& 0.670 & 0.608 & 0.596
& 0.701 & 0.592 & 0.561
& 0.719 & 0.558 & 0.535\\
\bottomrule
\end{tabular}%
}
\vspace{-1em}
\end{table}

We evaluate verification accuracy on rollouts with commitment decisions determined by the oracle criterion at $\epsilon=0.24$. Each chunk is denoised through all four steps to obtain complete convergence labels, while its output is assembled from the estimates selected by oracle commitment and used as history for subsequent chunks. Table~\ref{tab:sup_accuracy} reports AUROC across three labeling tolerances, with the rollout commitment tolerance fixed at $\epsilon=0.24$. We compute AUROC within each chunk and average over chunks, since at run time the verifier only ranks tokens inside the chunk it is about to denoise.

A per-token MLP achieves verification accuracy comparable to Ours\textsubscript{H}, suggesting that the backbone's contextualized token features already provide useful convergence cues. Training on-policy, where the head sees the truncated history during training as well, is no better even though these measurements are taken on that distribution. 

The latent-based signals support our training-free design. Redundancy is available at the first step but becomes less useful as a token is refined, while stability is stronger afterwards yet undefined at the first step. Our training-free verifier therefore uses redundancy at the first step and stability after, at the cost of two thresholds. Motion-based scheduling stays below redundancy at every step and tolerance. The head leads every latent signal throughout.

\subsection{Additional Ablation on Our Head}
\label{app:additional_ablation}

\begin{table}[t]
\centering
\footnotesize
\tabcolsep=0.8em
\renewcommand{\arraystretch}{1.2}
\caption{\textbf{Verifier ablations on Lingbot-World 2.0.} Learned verifier variants are calibrated to our operating point ($\epsilon=0.24$).}
\label{tab:sup_abl}
\vspace{-0.5em}
\begin{tabular}{l c c c c c}
\toprule
Method &
\makecell{Mean\\Steps $\downarrow$} &
\makecell{Through.\\(FPS)} &
\makecell{Drift\\$\downarrow$} &
\makecell{Quality\\Score $\uparrow$} &
\makecell{Imaging\\Quality $\uparrow$}\\
\midrule
On-policy training & 2.57 & 24.5 & 2.940 & \textbf{82.67} & 72.40 \\
Per-token MLP & 2.52 & 24.8 & 2.899 & \textbf{82.67} & 72.48 \\
w/o spatiotemporal smoothing & 2.55 & 24.7 & \underline{2.882} & 82.58 & \underline{72.54} \\
\textbf{Ours\textsubscript{H}} & 2.55 & 24.5 & \textbf{2.839} & \underline{82.65} & \textbf{72.65} \\
\bottomrule
\end{tabular}
\vspace{-1em}
\end{table}

We evaluate head variants under the same Lingbot-World 2.0 benchmark setting, with thresholds calibrated to obtain comparable mean denoising steps. As shown in Table~\ref{tab:sup_abl}, training on commitment rollouts or replacing the convolutional head with a per-token MLP yields similar generation quality and throughput. Spatiotemporal smoothing encourages coherent acceptance decisions across neighboring tokens and achieves lower Drift and higher imaging quality, supporting more stable rollouts.

\subsection{Additional Quantitative Results}
\label{app:quantitative_results}

\begin{table}[t]
\centering
\caption{Quality evaluation on LongLive. Evaluated on VBench.}
\label{tab:sup_5s_qual}
\resizebox{\textwidth}{!}{%
\begin{tabular}{l c c c c c c c c c c c}
\toprule
Method & \makecell{Mean\\Steps} & \makecell{Through.\\(FPS)} & Total & Quality & \makecell{Subject\\Consist.} & \makecell{Backgr.\\Consist.} & \makecell{Temporal\\Flicker} & \makecell{Motion\\Smooth.} & \makecell{Dynamic\\Degree} & \makecell{Aesthetic\\Quality} & \makecell{Imaging\\Quality} \\
\midrule
1-step & 1.00 & 67.97 & 79.06 & 80.03 & 90.43 & 93.51 & 99.20 & 98.24 & 45.00 & 61.06 & 63.97 \\
2-step & 2.00 & 44.54 & 82.46 & 83.06 & 95.88 & 95.84 & 99.35 & 98.59 & 46.67 & 65.05 & 67.65 \\
3-step & 3.00 & 33.15 & 83.08 & 83.45 & 96.91 & 96.75 & 99.40 & 98.76 & 41.11 & 66.55 & 68.39 \\
4-step & 4.00 & 26.40 & 83.08 & 83.49 & 97.13 & 96.98 & 99.38 & 98.84 & 37.78 & 67.21 & 68.82 \\
Surprise Forcing~\citep{surpriseforcing} & 3.59 & 27.38 & 83.11 & 83.53 & 97.13 & 96.99 & 99.39 & 98.83 & 38.33 & 67.23 & 68.80 \\
\textbf{Ours}$_{\mathrm{TF}}$ (matched to $\varepsilon{=}0.24$) & 1.97 & 39.77 & 83.16 & 83.55 & 96.97 & 96.51 & 99.43 & 98.76 & 41.67 & 66.64 & 68.83 \\
\textbf{Ours}$_{\mathrm{TF}}$ (matched to $\varepsilon{=}0.28$) & 1.65 & 43.61 & 83.12 & 83.51 & 96.64 & 96.34 & 99.41 & 98.67 & 44.17 & 66.26 & 68.62 \\
\textbf{Ours}$_{\mathrm{H}}$ ($\varepsilon{=}0.24$) & 1.83 & 39.93 & 83.20 & 83.58 & 96.91 & 96.47 & 99.42 & 98.74 & 42.22 & 66.56 & 69.03 \\
\textbf{Ours}$_{\mathrm{H}}$ ($\varepsilon{=}0.28$) & 1.68 & 41.74 & 83.21 & 83.54 & 96.77 & 96.30 & 99.41 & 98.70 & 43.61 & 66.33 & 68.88 \\
\bottomrule
\end{tabular}}
\end{table}

\begin{table}[t]
\centering
\caption{Semantic evaluation on LongLive. Evaluated on VBench.}
\label{tab:sup_5s_sem}
\resizebox{\textwidth}{!}{%
\begin{tabular}{l c c c c c c c c c c c c c}
\toprule
Method & \makecell{Mean\\Steps} & \makecell{Through.\\(FPS)} & Total & Semantic & \makecell{Object\\Class} & \makecell{Multiple\\Objects} & \makecell{Human\\Action} & \makecell{Colour} & \makecell{Spatial\\Relation} & \makecell{Scene} & \makecell{Appear.\\Style} & \makecell{Temporal\\Style} & \makecell{Overall\\Consist.} \\
\midrule
1-step & 1.00 & 67.97 & 79.06 & 75.17 & 90.27 & 64.57 & 94.80 & 91.03 & 68.39 & 46.45 & 21.11 & 23.28 & 26.62 \\
2-step & 2.00 & 44.54 & 82.46 & 80.09 & 95.52 & 79.88 & 97.20 & 90.92 & 77.32 & 55.77 & 20.62 & 24.06 & 26.92 \\
3-step & 3.00 & 33.15 & 83.08 & 81.58 & 96.80 & 87.20 & 96.60 & 91.25 & 78.87 & 58.75 & 20.58 & 24.20 & 26.78 \\
4-step & 4.00 & 26.40 & 83.08 & 81.43 & 96.27 & 87.26 & 96.20 & 90.78 & 78.21 & 59.65 & 20.55 & 24.21 & 26.65 \\
Surprise Forcing~\citep{surpriseforcing} & 3.59 & 27.38 & 83.11 & 81.42 & 96.28 & 87.18 & 96.20 & 90.87 & 78.22 & 59.52 & 20.55 & 24.21 & 26.66 \\
\textbf{Ours}$_{\mathrm{TF}}$ (matched to $\varepsilon{=}0.24$) & 1.97 & 39.77 & 83.16 & 81.62 & 96.49 & 86.23 & 97.40 & 91.04 & 79.47 & 59.46 & 20.47 & 24.23 & 26.72 \\
\textbf{Ours}$_{\mathrm{TF}}$ (matched to $\varepsilon{=}0.28$) & 1.65 & 43.61 & 83.12 & 81.56 & 96.36 & 86.60 & 97.60 & 91.16 & 79.57 & 58.42 & 20.42 & 24.28 & 26.77 \\
\textbf{Ours}$_{\mathrm{H}}$ ($\varepsilon{=}0.24$) & 1.83 & 39.93 & 83.20 & 81.68 & 96.30 & 86.97 & 97.60 & 90.65 & 80.32 & 58.90 & 20.47 & 24.26 & 26.73 \\
\textbf{Ours}$_{\mathrm{H}}$ ($\varepsilon{=}0.28$) & 1.68 & 41.74 & 83.21 & 81.89 & 96.60 & 88.49 & 97.40 & 91.41 & 80.23 & 58.58 & 20.43 & 24.27 & 26.76 \\
\bottomrule
\end{tabular}}
\end{table}

\begin{table}[t]
\centering
\caption{Quality evaluation on LongLive. Evaluated on VBench-Long.}
\label{tab:sup_30s_qual}
\resizebox{\textwidth}{!}{%
\begin{tabular}{l c c c c c c c c c c c c}
\toprule
Method & \makecell{Mean\\Steps} & \makecell{Through.\\(FPS)} & Drift & Total & Quality & \makecell{Subject\\Consist.} & \makecell{Backgr.\\Consist.} & \makecell{Temporal\\Flicker} & \makecell{Motion\\Smooth.} & \makecell{Dynamic\\Degree} & \makecell{Aesthetic\\Quality} & \makecell{Imaging\\Quality} \\
\midrule
1-step & 1.00 & 61.59 & 2.930 & 80.82 & 81.87 & 96.94 & 95.76 & 99.24 & 98.24 & 49.63 & 60.09 & 63.81 \\
2-step & 2.00 & 40.91 & 1.992 & 82.32 & 82.99 & 97.67 & 96.38 & 99.27 & 98.60 & 44.81 & 62.92 & 67.66 \\
3-step & 3.00 & 30.62 & 1.738 & 82.60 & 83.09 & 98.09 & 96.75 & 99.31 & 98.77 & 38.98 & 63.71 & 68.69 \\
4-step & 4.00 & 24.45 & 1.722 & 82.69 & 83.29 & 98.22 & 96.99 & 99.28 & 98.85 & 39.17 & 64.43 & 68.52 \\
Surprise Forcing~\citep{surpriseforcing} & 3.22 & 27.61 & 1.674 & 82.89 & 83.52 & 98.19 & 97.04 & 99.35 & 98.80 & 41.57 & 64.80 & 68.44 \\
\textbf{Ours}$_{\mathrm{TF}}$ & 1.71 & 38.78 & 1.685 & 82.75 & 83.28 & 98.11 & 96.77 & 99.44 & 98.74 & 41.57 & 63.83 & 68.27 \\
\textbf{Ours}$_{\mathrm{H}}$ ($\varepsilon{=}0.24$) & 1.72 & 37.78 & 1.707 & 82.81 & 83.32 & 98.22 & 96.79 & 99.42 & 98.75 & 40.74 & 63.95 & 68.66 \\
\textbf{Ours}$_{\mathrm{H}}$ ($\varepsilon{=}0.28$) & 1.55 & 39.43 & 1.742 & 82.88 & 83.43 & 98.02 & 96.68 & 99.43 & 98.69 & 44.72 & 63.57 & 68.37 \\
\bottomrule
\end{tabular}}
\end{table}

\begin{table}[t]
\centering
\caption{Semantic evaluation on LongLive. Evaluated on VBench-Long.}
\label{tab:sup_30s_sem}
\resizebox{\textwidth}{!}{%
\begin{tabular}{l c c c c c c c c c c c c c c}
\toprule
Method & \makecell{Mean\\Steps} & \makecell{Through.\\(FPS)} & Drift & Total & Semantic & \makecell{Object\\Class} & \makecell{Multiple\\Objects} & \makecell{Human\\Action} & \makecell{Colour} & \makecell{Spatial\\Relation} & \makecell{Scene} & \makecell{Appear.\\Style} & \makecell{Temporal\\Style} & \makecell{Overall\\Consist.} \\
\midrule
1-step & 1.00 & 61.59 & 2.930 & 80.82 & 76.64 & 90.99 & 71.92 & 95.67 & 92.21 & 67.74 & 49.88 & 21.08 & 23.39 & 26.40 \\
2-step & 2.00 & 40.91 & 1.992 & 82.32 & 79.61 & 93.58 & 81.67 & 98.00 & 91.87 & 75.71 & 52.77 & 20.74 & 23.92 & 26.65 \\
3-step & 3.00 & 30.62 & 1.738 & 82.60 & 80.67 & 92.24 & 86.96 & 97.33 & 91.28 & 78.54 & 56.14 & 20.60 & 24.11 & 26.61 \\
4-step & 4.00 & 24.45 & 1.722 & 82.69 & 80.30 & 92.47 & 84.48 & 95.67 & 89.56 & 79.43 & 57.83 & 20.58 & 24.09 & 26.43 \\
Surprise Forcing~\citep{surpriseforcing} & 3.22 & 27.61 & 1.674 & 82.89 & 80.39 & 92.85 & 84.55 & 96.33 & 89.99 & 79.43 & 56.77 & 20.58 & 24.14 & 26.58 \\
\textbf{Ours}$_{\mathrm{TF}}$ & 1.71 & 38.78 & 1.685 & 82.75 & 80.65 & 92.75 & 84.93 & 97.33 & 90.09 & 79.74 & 57.76 & 20.55 & 23.91 & 26.63 \\
\textbf{Ours}$_{\mathrm{H}}$ ($\varepsilon{=}0.24$) & 1.72 & 37.78 & 1.707 & 82.81 & 80.77 & 92.46 & 85.57 & 97.00 & 90.77 & 78.68 & 58.95 & 20.53 & 23.97 & 26.60 \\
\textbf{Ours}$_{\mathrm{H}}$ ($\varepsilon{=}0.28$) & 1.55 & 39.43 & 1.742 & 82.88 & 80.68 & 92.26 & 84.81 & 97.67 & 91.06 & 79.25 & 57.81 & 20.52 & 23.95 & 26.62 \\
\bottomrule
\end{tabular}}
\end{table}

We report all quality and semantic dimensions for LongLive in Tables~\ref{tab:sup_5s_qual} and~\ref{tab:sup_5s_sem} for 5-second videos using VBench, and Tables~\ref{tab:sup_30s_qual} and~\ref{tab:sup_30s_sem} for 30-second videos using VBench-Long. We compare uniform sampling with one, two, three, and four denoising steps against Speculative Forcing at $\epsilon=0.24$ and $\epsilon=0.28$.

We additionally implement the step allocation policy of Surprise Forcing~\citep{surpriseforcing} on LongLive, retaining LongLive's original memory management. For each chunk, the policy ranks the maximum cosine distance between adjacent latent frames of the first clean estimate against the preceding 20 chunks. Chunks below the rank threshold of $0.4$ use only the first and fourth denoising steps, while the others follow the full schedule. We retain the published settings, including full-step generation for the first five chunks, which cover five of the seven chunks in 5-second videos.

\subsection{Additional Qualitative Results}
\label{app:qualitative_results}
Additional qualitative results are provided in \textbf{the supplementary video materials}.

\section{Discussions}
\label{app:discussions}

\subsection{Limitations}
\label{app:limitations}

The achievable acceleration depends on both the underlying generator and the difficulty of the generation domain. If the backbone is sensitive to reducing denoising steps, fewer tokens can be accepted early without degrading generation quality, which limits the attainable speedup. Similarly, more challenging domains may require additional refinement across a larger fraction of tokens. This is evident in world generation, where large viewpoint changes introduce stronger distortions than standard video generation and result in higher mean denoising steps, as shown in Table~\ref{tab:main_lingbot}. Our method therefore does not guarantee the same acceleration across different generators or domains.

\subsection{Other Discussions}
\label{app:other_discussions}

\paragrapht{Mean Steps and Realized Speed-up.}
The clean forward pass used to refresh the KV cache adds one full pass per chunk. Assuming computation scales linearly with token evaluations, the ideal speed-up relative to the four-step generator is $5/(\bar{k}+1)$. Approaches that reuse noisy KV states, such as CausalRCM~\citep{causalrcm}, could avoid this pass and move the ideal speed-up toward $4/\bar{k}$.

In practice, throughput does not scale proportionally with the mean denoising depth $\bar{k}$. Each chunk continues until its last active token is accepted, incurring forward-pass overhead at every executed step even when few tokens remain active. This overhead therefore depends on $\max_i k^\ast(i)$ within each chunk rather than the average across tokens. Reducing this overhead would bring the realized speed-up closer to the ideal implied by the mean denoising depth.

\subsection{Potential Applications}
\label{app:applications}

Our experiments focus on general video generation and world modeling, but token-wise commitment may be particularly effective when much of the target content is already constrained by available conditions. In talking-head generation~\citep{liveavatar}, for example, the background can remain largely unchanged while motion is concentrated around the face and upper body. In environment-grounded world models such as Seoul World Model~\citep{swm} and World Observer~\citep{worldobserver}, warped scene observations and reference images provide visual evidence for parts of the target view, while uncovered regions require new content to be generated. Dexterous World Models~\citep{dwm} provide static scene renderings along the camera trajectory and generate interaction-induced changes while preserving unaffected content.

These settings offer opportunities to allocate denoising computation according to how much refinement remains after conditioning. Tokens already reproducing content supported by scene conditions or references may be accepted early, while newly generated or dynamically changing regions receive further refinement. Extending token-wise commitment to these settings could also support efficient augmented and virtual reality applications that combine known environments with generated content and interaction dynamics.

\end{document}